\documentclass{article}

\usepackage{arxiv}

\usepackage{amsmath,amsfonts}
\usepackage{url}
\usepackage{orcidlink}
\usepackage{booktabs}
\usepackage{multirow}
\usepackage{makecell}

\newcommand{\acks}[1]{\section*{Acknowledgements}#1}
\newcommand{\ethics}[1]{\section*{Ethics Statement}#1}
\newcommand{\coi}[1]{\section*{Conflict of Interest}#1}
\newcommand{\data}[1]{\section*{Data Availability}#1}

\usepackage{tikz}
\usetikzlibrary{shapes.geometric, arrows, positioning}

\tikzstyle{process} = [rectangle, minimum width=3cm, minimum height=1cm, text centered, draw=black, fill=blue!30]
\tikzstyle{arrow} = [thick,->,>=stealth]

\title{DICOM De-Identification via Hybrid AI and Rule-Based Framework for Scalable, Uncertainty-Aware Redaction}

\author{%
  Kyle Naddeo\thanks{%
    ORCID: \href{https://orcid.org/0009-0005-1651-9733}{0009-0005-1651-9733};%
    \quad Email: \texttt{naddeok5@rowan.edu};%
    \quad Affiliation: Rowan University, Glassboro NJ 08028, USA%
  }%
  \and
  Nikolas Koutsoubis\thanks{%
    ORCID: \href{https://orcid.org/0000-0001-6195-9360}{0000-0001-6195-9360};%
    \quad Email: \texttt{niko.koutsoubis@moffitt.org};%
    \quad Affiliation: Moffitt Cancer Center, Tampa FL 33612, USA; University of South Florida, Tampa FL 33620, USA%
  }%
  \and
  Rahul Krish\thanks{%
    ORCID: \href{https://orcid.org/0009-0005-3820-6199}{0009-0005-3820-6199};%
    \quad Email: \texttt{rahul@ibisworks.com};%
    \quad Affiliation: Impact Business Information Solutions, Inc, Princeton NJ 08542, USA%
  }%
  \and
  Ghulam Rasool\thanks{%
    ORCID: \href{https://orcid.org/0000-0001-8551-0090}{0000-0001-8551-0090};%
    \quad Email: \texttt{ghulam.rasool@moffitt.org};%
    \quad Affiliation: Moffitt Cancer Center, Tampa FL 33612, USA; University of South Florida, Tampa FL 33620, USA%
  }%
  \and
  Nidhal Bouaynaya\thanks{%
    ORCID: \href{https://orcid.org/0000-0002-8833-8414}{0000-0002-8833-8414};%
    \quad Email: \texttt{bouaynaya@rowan.edu};%
    \quad Affiliation: Rowan University, Glassboro NJ 08028, USA%
  }%
  \and
  Tony O’Sullivan\thanks{%
    ORCID: \href{https://orcid.org/0000-0001-8277-9115}{0000-0001-8277-9115};%
    \quad Email: \texttt{tosullivan@ibisworks.com};%
    \quad Affiliation: Impact Business Information Solutions, Inc, Princeton NJ 08542, USA%
  }%
  \and
  Raj Krish\thanks{%
    ORCID: \href{https://orcid.org/0009-0004-9069-2414}{0009-0004-9069-2414};%
    \quad Email: \texttt{rajkrish@ibisworks.com};%
    \quad Affiliation: Impact Business Information Solutions, Inc, Princeton NJ 08542, USA%
  }%
}

\begin{document}

\maketitle

\begin{abstract}
Access to medical imaging and associated text data has the potential to drive major advances in healthcare research and patient outcomes. However, the presence of Protected Health Information (PHI) and Personally Identifiable Information (PII) in Digital Imaging and Communications in Medicine (DICOM) files presents a significant barrier to the ethical and secure sharing of imaging datasets. This paper presents a hybrid de-identification framework developed by Impact Business Information Solutions (IBIS) that combines rule-based and AI-driven techniques, and rigorous uncertainty quantification for comprehensive PHI/PII removal from both metadata and pixel data.

Our approach begins with a two-tiered rule-based system targeting explicit and inferred metadata elements, further augmented by a large language model (LLM) fine-tuned for Named Entity Recognition (NER), and trained on a suite of synthetic datasets simulating realistic clinical PHI/PII. For pixel data, we employ an uncertainty-aware Faster R-CNN model to localize embedded text, extract candidate PHI via Optical Character Recognition (OCR), and apply the NER pipeline for final redaction. Crucially, uncertainty quantification provides confidence measures for AI-based detections to enhance automation reliability and enable informed human-in-the-loop verification to manage residual risks.

This uncertainty-aware deidentification framework achieves robust performance across benchmark datasets and regulatory standards, including DICOM, HIPAA, and TCIA compliance metrics. By combining scalable automation, uncertainty quantification, and rigorous quality assurance, our solution addresses critical challenges in medical data de-identification and supports the secure, ethical, and trustworthy release of imaging data for research.
\end{abstract}


\section{Introduction}
Ensuring patient privacy while enabling the secondary use of medical imaging data remains a fundamental challenge in modern biomedical research \cite{iom09, shojaei2024security}. Digital Imaging and Communications in Medicine (DICOM) files often include Protected Health Information (PHI) and Personally Identifiable Information (PII) embedded both in structured metadata fields and as ``burned-in'' text within the image pixel data. Any reuse or public release of such data requires robust de-identification protocols that are compliant with regulatory standards like the Health Insurance Portability and Accountability Act (HIPAA) in the United States and the General Data Protection Regulation (GDPR) in Europe \cite{freymann12,aryanto16,kayode2024}.

Traditional de-identification methods have relied heavily on manual data masking or software tools designed to remove identifiable information from DICOM files \cite{ocr23,clunie2023midi}. While these approaches can be effective, they are labor-intensive, slow, and prone to human error \cite{ovic2024}. Recent advancements have led to automated de-identification systems leveraging natural language processing (NLP), computer vision (CV), and machine learning (ML) to improve accuracy and scalability, reduce manual burden, and efficiently detect and redact PHI and PII while preserving clinical data utility \cite{shahid2022,moore2015,diaz2021data,murugadoss21,chambon2023}.

The integration of artificial intelligence (AI) and machine learning (ML) techniques has notably improved de-identification performance. Deep CNN-based methods,  including object detection algorithms such as YOLO and Faster R-CNN variants, automatically localize and redact textual identifiers embedded in medical images with high accuracy, surpassing manual processes in efficiency and adaptability across various imaging modalities \cite{Monteiro2017,Khosravi2023}. Meanwhile, transformer-based architectures, although less common in production settings, offer promising results due to their ability to model global image contexts more effectively \cite{He2023}. In practice, a hybrid workflow is adopted, combining rule-based metadata cleansing with deep learning models for pixel-level PHI removal. This hybrid strategy leverages established best practices to handle explicit fields while entrusting modern AI with subtle or burned-in text, thus maximizing both privacy protection and retention of relevant diagnostic information.

However, the effectiveness of AI-based systems is significantly hindered by the lack of available training data and proper uncertainty quantification \cite{ahmed2023,abdar2021,challen2019}. Medical image datasets often lack the diversity and volume necessary to train robust AI models, primarily due to privacy concerns and the complexities involved in creating de-identified datasets. This scarcity of data limits the ability of AI systems to generalize across different imaging modalities and clinical contexts \cite{seastedt2022}. Moreover, current AI de-identification models often lack mechanisms to quantify and communicate their uncertainty in the predictions, making it difficult to gauge the reliability of model outputs. Without proper uncertainty estimation, there is an elevated risk of either inadvertently exposing PHI or unnecessarily removing clinically relevant information.

In response, our approach integrates novel uncertainty-aware ML models for both imaging and text data with traditional rule-based de-identification methods. The process begins with a rule-based system designed to eliminate explicit identifiers within DICOM metadata.  This initial step is enhanced by a fine-tuned large language model (LLM) for Named Entity Recognition (NER), trained using synthetic datasets crafted to mirror realistic clinical narratives containing diverse PHI scenarios. This synthetic data generation addresses the challenges of data scarcity and enhances the robustness and generalization of our models. 

For pixel data, we develop an uncertainty-aware Faster R-CNN model explicitly designed to identify and extract embedded textual regions from medical images. The model provides quantifiable measures of uncertainty and reliability for each detection. Once text regions are detected, optical character recognition (OCR) is employed to extract the text from these areas. This extracted text is then processed using the same robust text-based analysis framework applied to metadata, ensuring consistency and thoroughness in de-identification efforts. 

\section{Methods}
\subsection{Synthetic Data Generation}
\subsubsection{Metadata}
To simulate realistic clinical metadata for model training, we utilized the Python library Faker \cite{faker19} to populate DICOM tags with synthetic PHI, including names, dates, and addresses. In order to ensure that our synthetic metadata accurately reflects the structure and complexity of real-world medical data, we incorporated the standardized DICOM templates provided by the DICOM Standard \cite{DICOMStandard}, implementing data on a per-header basis. This approach enabled us to generate metadata that adheres to the specific header configurations typical in clinical imaging data, without using any actual PHI. Such measures are essential given strict privacy regulations, such as the Health Insurance Portability and Accountability Act (HIPAA) in the United States, which prohibit the use of identifiable health information in research unless specific exemptions are met, and the risks posed by reverse machine learning techniques.

Reverse machine learning refers to a set of techniques through which adversaries may attempt to reconstruct input data or sensitive patterns embedded in model weights by exploiting the parameters or gradients of trained models \cite{fredrikson15}. For instance, membership inference attacks \cite{shokri17} could determine whether specific data points were part of the training dataset, while model inversion attacks \cite{fredrikson15} might recreate identifiable features, such as names or dates, based on the model's learned parameters. These vulnerabilities pose significant risks when models are trained on real PHI, as they could inadvertently expose sensitive information even when the data itself is not directly shared.

By utilizing synthetic metadata generated by Faker, we effectively mitigated these risks. The synthetic data has no direct correlation to real individuals, thereby safeguarding against the reconstruction of identifiable information. This approach aligns with ethical research practices, facilitates compliance with data protection laws, and supports the development and testing of robust machine learning models designed for medical applications.

\subsubsection{Synthetic Admission Notes}
In this study, we first sampled hospital admission notes from the i2b2 dataset~\cite{i2b2_mayo} to serve as a ground truth dataset. These real clinical notes provided realistic content and structure for the prompt engineering of synthetic data generation. Using this ground truth, we generated synthetic hospital admission notes with ChatGPT-4~\cite{ChatGPT}. The synthetic notes were created by providing detailed instructions to ChatGPT-4 to produce structured admission notes conforming to a predefined clinical format. Each note included sections for Record Date, Admission Team, Patient Information (Name, Medical Record Number, Date of Birth), Primary Care Physician, Chief Complaint (CC) / Reason for Admission (RFA), History of Present Illness (HPI), Past Medical History (PMHx), Allergies, Medications, Social History (SHx), Family History (FHx), Physical Exam, Laboratory Results (formatted as a properly spaced table), Imaging Studies, Electrocardiogram (EKG), and Assessment \& Plan (A/P). Additionally, each note contained embedded protected health information (PHI) labels, such as AGE, DATE, EMAIL, HOSPITAL, ID, LOCATION, OTHERPHI, PATIENT, PATORG, PHONE, and STAFF. ChatGPT-4 was further instructed to produce a corresponding Python dictionary mapping named entity recognition (NER) labels to their respective start and end indices within the generated text. Notably, all synthetic notes were generated solely through prompt engineering without additional model fine-tuning. A total of 1,532 synthetic notes were generated, encompassing 33,750 labeled entities, which serve as training data for fine-tuning our NER models.

\subsubsection{Pixel Data}\label{sec: data for pixeldata}
To create synthetic pixel data, we first obtained clean medical images from the Cancer Imaging Archive \cite{clark13}. Using Faker \cite{faker19}, we generated synthetic text data, some of which was classified as PHI, and embedded it into the images at random locations using OpenCV \cite{opencv2015}. This included adding text onto objects within the images, such as jewelry, and varying font styles and sizes to simulate realistic scenarios, including burned-in annotations.

A key aspect of our approach was distinguishing PHI from non-PHI text, as not all textual information in medical images meets the criteria for protected health information under regulations like HIPAA. For example, general annotations such as imaging parameters (e.g., "AXIAL T2") or anatomical markers ("LEFT," "RIGHT") are not PHI but can be critical for downstream tasks like quality control or research. Removing all text indiscriminately would unnecessarily degrade the utility of the images.

To address this, our system was designed to selectively identify and redact only PHI, preserving non-PHI information whenever possible. This targeted approach reflects our guiding principle: ``do the least amount of damage while removing all PHI." The utility of the dataset for potential downstream tasks is therefore maximized by preserving non-PHI text, while fully complying with privacy requirements. This method ensures privacy is preserved without compromising the broader applicability of the data, and maintains essential contextual information that could otherwise be lost through blanket redaction.

The generated dataset consisted of 10{,}000 synthetic images, of which 8{,}000 were used for training, while the remaining 2{,}000 were split equally between validation and testing. To further enhance the diversity of the dataset, we applied several augmentation techniques, including \emph{ResizeShortestEdge}, \emph{RandomFlip}, and salt-and-pepper noise to expose the model to a wide variety of potential real-world conditions.

We also ensured that the synthetic data represented different imaging modalities, including CT, MRI, and X-ray, which were equally balanced across the dataset. This comprehensive approach allowed us to train our models effectively, despite the absence of real-world PHI during the training phase.

Lastly, while the training was conducted exclusively on synthetic data, the models were validated using real-world medical DICOMS obtained from the Moffitt Cancer Center. This validation step was crucial to confirm the reliability and accuracy of our de-identification framework.

\subsection{Algorithmic Approach}
\begin{figure}[ht]
    \centering
    \includegraphics[width=1\linewidth]{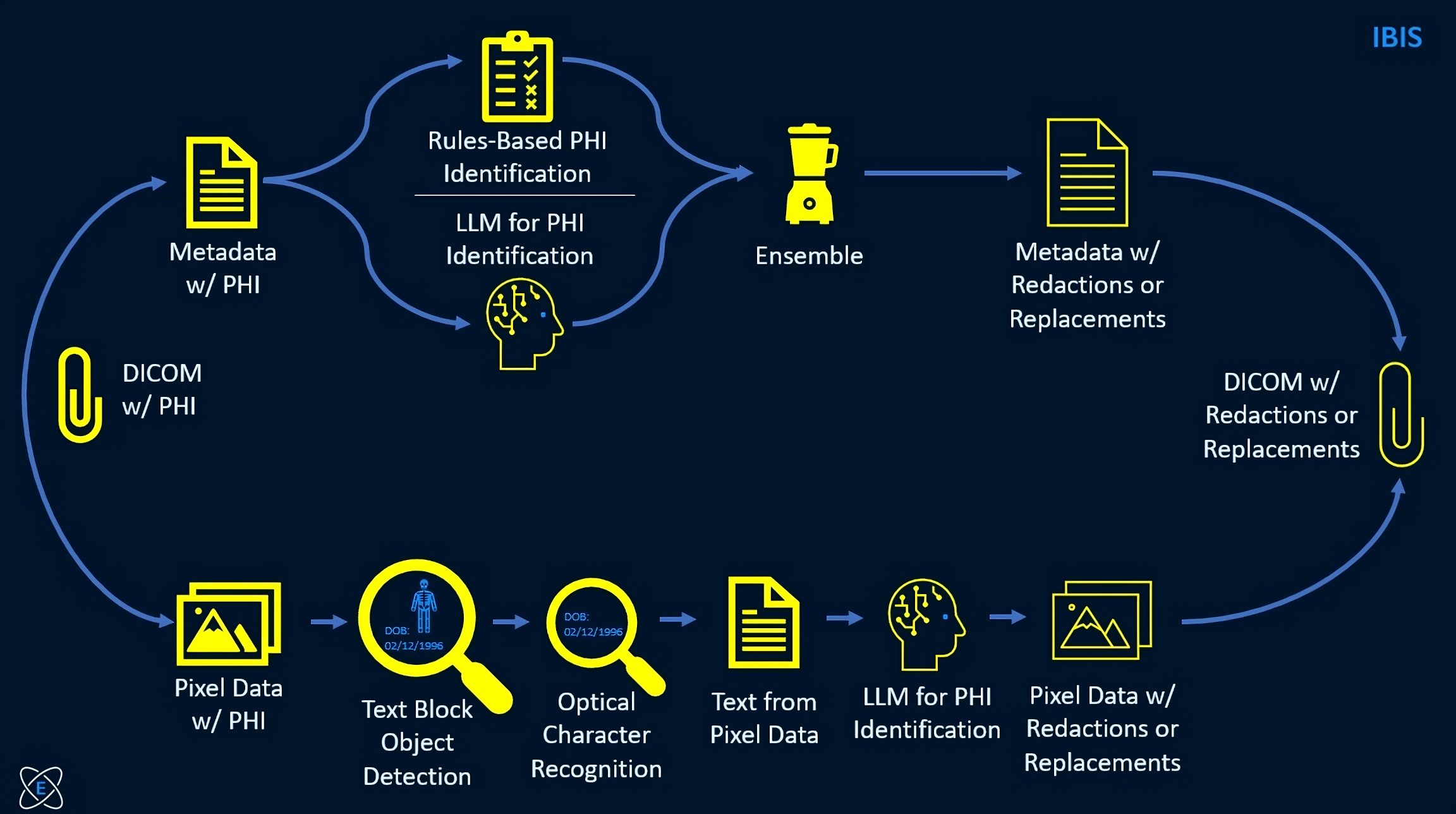}
    
    \caption{This diagram illustrates our general workflow for DICOM-based deidentification. A DICOM file consists of two main components: a header containing metadata and pixel data representing the image. For metadata, an ensemble of rule-based methods and natural language processing (NLP) is used to identify and redact Protected Health Information (PHI). For pixel data, object detectors isolate text regions, Optical Character Recognition (OCR) translates the text from pixel space, and NLP identifies PHI to be redacted.}
    
    \label{fig:deid_workflow}
\end{figure}

Our de-identification framework, illustrated in Figure~\ref{fig:deid_workflow}, employs a dual-path approach to process both metadata and pixel data in DICOM images, ensuring thorough removal of Protected Health Information (PHI) and Personally Identifiable Information (PII).

\subsubsection{Metadata}
The text metadata processing begins with a rule-based system that targets specific DICOM tags known to contain PHI. This system, which we refer to as “recipes,” identifies and removes fields based on predefined rules. After the initial removal, the extracted data is further processed using Python’s \emph{FuzzyWuzzy} package \cite{fuzzy21}. This package leverages Levenshtein Distance to calculate the differences between sequences, allowing us to detect and remove subtle identifiers that may be spread throughout the text (see Figure~\ref{fig:levenshtein_example}). Notably, if ``John" is flagged as PHI by our system, then ``Jon" (or any other string within the threshold of Levenshtein Distance) will also be recognized as PHI.

\begin{figure}[h!]
\centering
\begin{tabular}{p{0.45\textwidth} p{0.45\textwidth}}
\toprule
\multicolumn{1}{c}{\textbf{Example 1}} & \multicolumn{1}{c}{\textbf{Example 2}}\\
\midrule
\textbf{DICOM Tag 1:} "PatientName: John Doe" \newline
\textbf{DICOM Tag 2:} "PatientName: Jon Doe" \newline
Levenshtein Distance: 1 \newline
(`h` is removed from "John" to become "Jon") 
&
\textbf{DICOM Tag 1:} "PatientName: Jane Smith" \newline
\textbf{DICOM Tag 2:} "PatientName: Bugs Bunny" \newline
Levenshtein Distance: 9 \newline
(replace `J` with `B`, `a` with `u`, `n` with `g`, `e` with `s`, and "Smith" with "Bunny")
\\
\bottomrule
\end{tabular}
\caption{Illustration of how Levenshtein Distance quantifies textual differences in DICOM tags. In Example 1, the difference is minimal (distance = 1). In Example 2, the distance is significantly larger (distance = 9), reflecting more substantial changes. By setting an appropriate threshold on the Levenshtein Distance, our system treats near matches (e.g., “John” vs. “Jon”) as the same PHI and therefore de-identifies all variants.}
\label{fig:levenshtein_example}
\end{figure}

To enhance the rule-based process, we incorporate the pre-trained named entity language model LUKE \cite{luke20}, originally trained on tasks such as entity typing, relation classification, and named entity recognition across datasets like Open Entity \cite{choi2018}, TACRED \cite{zhang2017}, and CoNLL-2003 \cite{tjong2003}. We then fine-tuned this model on our synthetic hospital admission notes using default hyperparameters, which effectively equipped the model to detect PHI in longer clinical narratives.

However, we discovered that fine-tuning exclusively on admission notes did not adequately prepare the model for DICOM metadata fields, where PHI often appears in short, context-poor snippets. To address this gap, we conducted an additional fine-tuning step on synthetic metadata, pairing each DICOM tag with its corresponding value. This explicit ``tag-with-value" setup added crucial context and significantly improved the models’ ability to accurately identify short or single-word PHI. 

Finally, the ensemble system integrates the outputs of these AI models with the rule-based methods by taking the union of their detections, ensuring comprehensive de-identification across both longer narratives and discrete metadata fields.

\subsubsection{Pixel Data}\label{sec: algo for pixeldata}

Our approach employs an uncertainty-aware Faster R-CNN \cite{ren2016,dera2021premium} model to detect regions within images that may contain PHI or PII, such as embedded textual information. The detection framework is convolutional-based, consisting of a series of convolutional layers for feature extraction that feed into dense decision layers. In Faster R-CNN, the network operates in two stages. In the first stage, a Region Proposal Network (RPN) generates candidate regions by associating each anchor (a predefined template bounding box size based on training data statistics) with an objectness score and bounding-box regression parameters. For each anchor \(i\), let \(p_i\) denote the predicted probability of containing an object (1 for object, 0 for background), and \(t_i = (t_{x,i}, t_{y,i}, t_{w,i}, t_{h,i}) \)  represents the predicted bounding-box adjustment relative to an anchor box. Anchor boxes are predefined reference bounding boxes used to simplify and enhance the detection process by providing initial estimates of object shapes and sizes. The adjustments are defined as follows:
\begin{equation}
t_{x,i} = \frac{x_i - x_a}{w_a}, \quad t_{y,i} = \frac{y_i - y_a}{h_a}, \quad t_{w,i} = \log\left(\frac{w_i}{w_a}\right), \quad t_{h,i} = \log\left(\frac{h_i}{h_a}\right).
\end{equation}
where \((x_a, y_a, w_a, h_a)\) denote the parameters of the anchor box, and \((x_i, y_i, w_i, h_i)\) represent the coordinates of the actual predicted bounding box. In the second stage, the network classifies these region proposals and further refines their bounding box parameters.
The overall loss is given by 
\begin{equation}\label{eq: original rcnn loss}
L(p_i, \{t_i\}) = \frac{1}{N_{cls}} \sum_i L_{cls}(p_i, p_i^*) + \lambda\, \frac{1}{N_{reg}} \sum_i p_i^*\, L_{reg}(t_i, t_i^*),
\end{equation}
where \(p_i^*\) is the ground-truth probability and \(t_i^* \) is the target adjustment, i.e., adjustment with respect to the ground-truth coordinates. 

In order to quantify model uncertainty and improve Human interpretability of the outputs, we adopt our prior work on Bayesian variational inference formulation in which the inputs $x$, weights $W$, and biases $b$ are modeled as random variables  \cite{dera2021premium}. 
 
Consider a fully connected Bayesian neural network $W\in \mathbb{R}^{m\times d} = [\boldsymbol{w}_1, \ldots, \boldsymbol{w}_d]$, where $\boldsymbol{w}_i\in\mathbb{R}^{m}$ is the $i^{th}$ column of $W$ with mean $\boldsymbol{\mu}_{w_i}\in\mathbb{R}^{m}$ and an associated covariance matrix $\Sigma_{w_i}\in\mathbb{R}^{m\times m}$ and its diagonal elements, the variance vector, is $ \boldsymbol{\sigma}_{w_i}^2\in\mathbb{R}^{m} = \text{diag} (\Sigma_{w_i})$. The weight parameters follow a distribution such that
\begin{equation}
W \sim \mathcal{N}( \mbox{M}_W , \Sigma_W),
\end{equation}
where \(\Sigma_W\in\mathbb{R}^{d\times m \times m} = \text{Cov} (\text{vec} (W))\) is the covariance tensor of the vectorized random matrix.
The input \(\boldsymbol{x} \in \mathbb{R}^d\) and the bias vector \(\boldsymbol{b} \in \mathbb{R}^m\) are also assumed stochastic, 
\begin{eqnarray}
\boldsymbol{x} \sim \mathcal{N}(\boldsymbol{\mu}_x, \Sigma_x), \\
\boldsymbol{b} \sim \mathcal{N}(\boldsymbol{\mu}_b, \Sigma_b).
\end{eqnarray}

Under a linear transformation , we have 
\begin{equation}
\boldsymbol{z} = W\boldsymbol{x} + \boldsymbol{b},
\end{equation}
the output mean $\boldsymbol{\mu}_z \in \mathbb{R}^m$ is obtained as 
\begin{equation}
\boldsymbol{\mu}_z = \mbox{M}_W\,\boldsymbol{\mu}_x + \boldsymbol{\mu}_b,
\end{equation}
and the full covariance $\Sigma_z \in \mathbb{R}^{m\times m}$ is given for each element $i,j = 1,\ldots,m$ by 

\begin{equation}
\Sigma_{z,ij} =
\begin{cases}
\displaystyle
\operatorname{tr}\!\bigl(\Sigma_{w_i}\Sigma_{b}\bigr)
\;+\;
\boldsymbol{\mu}_{w_i}^{\!\top}\Sigma_{b}\,\boldsymbol{\mu}_{w_j}
\;+\;
\boldsymbol{\mu}_{b}^{\!\top}\Sigma_{w_i}\,\boldsymbol{\mu}_{b},
& \text{if } i = j,\\[8pt]
\displaystyle
\boldsymbol{\mu}_{w_i}^{\!\top}\Sigma_{b}\,\boldsymbol{\mu}_{w_j},
& \text{if } i \neq j.
\end{cases}
\end{equation}


While full covariance propagation captures correlations among the output dimensions, it is computationally expensive, particularly in deep architectures. For computational efficiency, especially in the dense decision head, we simplify the propagation by retaining only the diagonal elements of the covariance matrices, thereby propagating uncertainty solely in terms of variances. In this variance-only formulation, let \(\boldsymbol{\sigma}_x^2 \in \mathbb{R}^d\) denote the variance vector of \(x\) or equivalently $\boldsymbol{\sigma}_x^2 = \text{diag}(\Sigma_x)$ is the vector containing the diagonal elements of $\Sigma_x$. 
Thus, the mean and variance elements, $\mu_{z_i}$ and $\sigma^{2}_{z_i}$, of the random vector $\boldsymbol{z}$ can be derived for each element $i=1, \ldots, m$ as follows: 
\begin{equation}\label{eq:prop}
    \begin{split}
    \mu_{z_{i}}\!\!&= \boldsymbol{\mu}_{w_i} \cdot \boldsymbol{\mu}_{x} + \mu_{b_{i}}, \text{ and} \\
    \sigma^2_{z_{i}}\!\!&=
    (\boldsymbol{\mu}_{x}\odot\boldsymbol{\mu}_{x}) \cdot \boldsymbol{\sigma}^2_{w_i}
    + (\boldsymbol{\mu}_{w_i}\odot\boldsymbol{\mu}_{w_i}) \cdot \boldsymbol{\sigma}^2_{x}+{\sigma^2_{b_{i}}}, \\
    \end{split}
\end{equation}
where $\odot$ represents the element-wise product between two vectors.

When the output \(z\) is passed through a nonlinear activation function \(f\) (e.g., ReLU), a first-order Taylor expansion about \(\mu_z\) is employed,
\begin{equation}
f(z) \approx f(\boldsymbol{\mu}_z) + f'(\boldsymbol{\mu}_z)(\boldsymbol{z}-\boldsymbol{\mu}_z),
\end{equation}
where $f'(\boldsymbol{\mu}_z)\in\mathbb{R}^{d\times d}$ is the Jacobian and we consider second order and above terms neglible.  The output mean can now be defined as  
\begin{equation}
\boldsymbol{\mu}_a \approx f(\boldsymbol{\mu}_z)
\end{equation}
and the full covariance $\Sigma_a  \in \mathbb{R}^{d \times d}$ as
\begin{equation}
\Sigma_a \approx f'(\boldsymbol{\mu}_z)\,\Sigma_z\,f'(\boldsymbol{\mu}_z)^T.
\end{equation}
Following the variance-only assumption, the propagated variance simplifies to 
\begin{equation}
\boldsymbol{\sigma}_a^2 \approx (f'(\boldsymbol{\mu}_z)\odot f'(\boldsymbol{\mu}_z)) \cdot \boldsymbol{\sigma}_z^2.
\end{equation}
Thus, the network ultimately produces a vector of class probabilities, \(\boldsymbol{\mu}_{\text{probs}}\), along with an associated uncertainty measure that may be represented either as a full covariance matrix \(\Sigma_{\text{probs}}\) or, for efficiency, just the diagonal elements of \(\Sigma_{\text{probs}}\) as a variance vector \(\boldsymbol{\sigma}_{\text{probs}}^2\). The loss function comprises a negative log-likelihood (NLL) term, which, under the assumption that the network output follows a Gaussian distribution \(\mathcal{N}(\mbox{M}_y, \Sigma_y)\), is given by 
\begin{equation}
\mathcal{L}_{\text{NLL}} = \frac{1}{2}\Big[(\boldsymbol{y}-\boldsymbol{\mu}_y)^T\Sigma_y^{-1}(\boldsymbol{y}-\boldsymbol{\mu}_y) + \log\det\Sigma_y + k\log(2\pi)\Big],
\end{equation}
where \(k\) is the output dimensionality; in the variance-only case, this reduces to 
\begin{equation}\label{eq: var only nll}
\mathcal{L}_{\text{NLL}} \propto \frac{1}{2}\left[\frac{(\boldsymbol{y}-\boldsymbol{\mu}_y)^2}{\boldsymbol{\sigma}_y^2} + \log\boldsymbol{\sigma}_y^2\right].
\end{equation}
Additionally, a Kullback-Leibler (KL) divergence term is incorporated to regularize the approximate posterior distributions of the weights and biases against their priors. For example, if the prior for the weights is \(p(\boldsymbol{w}_i)=\mathcal{N}(0,I)\) and the approximate posterior is \(q(\boldsymbol{w}_i)=\mathcal{N}(\boldsymbol{\mu}_{w_i},\Sigma_{w_i})\), then the KL divergence is expressed as 
\begin{equation}
\text{KL}\bigl(q(\boldsymbol{w}_i)\parallel p(\boldsymbol{w}_i)\bigr) = \frac{1}{2}\Big[\operatorname{tr}(\Sigma_{w_i}) + \boldsymbol{\mu}_{w_i} \cdot \boldsymbol{\mu}_{w_i} - k - \log\det\Sigma_{w_i}\Big],
\end{equation}
where the constant \(k\) (reflecting the dimensionality of \(\boldsymbol{\mu}_{w_i}\)) may be omitted from the optimization since it does not affect the gradients. By removing constants and and adding coefficients the KL divergence loss for a layer can be defined as
\begin{equation}\label{eq: full kl loss}
\text{KL Layer Loss} = \frac{1}{\operatorname{ncols}W^{(l)}}\sum_{i}\Big[\lambda_1\operatorname{tr}(\Sigma_{w_i}) + \lambda_2 \boldsymbol{\mu}_{w_i} \cdot \boldsymbol{\mu}_{w_i} - \lambda_3 \log\det\Sigma_{w_i}\Big],
\end{equation}
As an initial proof of concept, we only replace the classification head of the network and thus modify only the first term in Equation~\ref{eq: original rcnn loss}, substituting it with the sum of the negative log-likelihood (Equation~\ref{eq: var only nll}) and the KL-divergence across all layers (Equation~\ref{eq: full kl loss}). We also introduce a coefficient for each term and remove non-optimizable factors such as constants and scaling. This leads to the following full loss:
\begin{align}
\mathcal{L}_{\text{total}}
= &\ \lambda_{\text{reg}}\, \frac{1}{N_{\text{reg}}} \sum_i p_i^*\, L_{\text{reg}}(t_i, t_i^*) \notag \\
&\ + \lambda_{\text{error over sigma}}\frac{(\boldsymbol{y}-\boldsymbol{\mu}_y)^2}{\boldsymbol{\sigma}_y^2} \;+\; \lambda_{\text{log det}} \log\boldsymbol{\sigma}_y^2 \notag \\
&\ + \sum_{l}\!\left[\frac{1}{\operatorname{ncols}W^{(l)}}\sum_{i}\Big[\lambda_{1,l}\operatorname{tr}(\Sigma_{w^{(l)}_i}) + \lambda_{2,l} \boldsymbol{\mu}_{w^{(l)}_i} \cdot \boldsymbol{\mu}_{w^{(l)}_i} - \lambda_{3,l} \log\det\Sigma_{w^{(l)}_i}\Big]\right] \label{eq:total_loss_line}
\end{align}
In this formulation, \(\lambda_{\text{reg}}\) governs the bounding-box regression accuracy, while \(\lambda_{\text{error over sigma}}\) and \(\lambda_{\text{log det}}\) balance the classification error term and the posterior covariance regularization for the predicted class probabilities, respectively. Finally, the sum of KL-divergence terms in each layer (weighted by \(\lambda_{1,l}\), \(\lambda_{2,l}\), and \(\lambda_{3,l}\)) aligns the weight distributions with a standard normal prior, effectively regularizing the weight covariance matrices. Consequently, the overall optimization framework provides a trade-off among bounding-box regression precision, classification accuracy, and posterior uncertainty management.

The integration of Variational Density Propagation \cite{dera2021premium} into the dense decision head not only produces point predictions but also quantifies the uncertainty associated with them. For our proof of concept, we analyze the uncertainty estimates only for final network predictions, ignoring regions with low confidence; however, there are cases where low confidence but high uncertainty may warrant further review, which we plan to explore in future work. We normalize the variance between the lowest and highest observed values, setting a practical threshold at the smallest normalized variance that yielded a false positive in testing. Text regions detected with variance below this threshold proceed to Tesseract OCR \cite{kay2007} for extraction and are subsequently de-identified using our hybrid rule-based and AI-assisted text analysis, while high-uncertainty detections are flagged for human-in-the-loop review.

To determine the optimal set of loss coefficients in Equation~\ref{eq:total_loss_line} (in our current experiment there are six), we defined five components for our objective function \(\Psi\). These components quantify different aspects of model performance and uncertainty estimation capabilities:
\begin{equation}
\Psi = \alpha \cdot \text{mAP}_{50} - \beta \cdot S_{\text{noise}} - \gamma \cdot V_{clean} - \delta \cdot \text{FNR}_{\text{thresh}} + \epsilon \cdot \text{IoU}_{\text{thresh}}
\label{eq:objective}
\end{equation}
where \(\alpha, \beta, \gamma, \delta, \epsilon\) are hyperparameters that control the relative importance of each term; all other elements will be defined in the remainder of this section.

The first component we use to evaluate our object detector is the mean Average Precision (mAP) at an Intersection-over-Union (IoU) threshold of 0.5, denoted mAP\(_{50}\). 

\begin{figure}[h]
    \centering
    \includegraphics[width=0.45\textwidth]{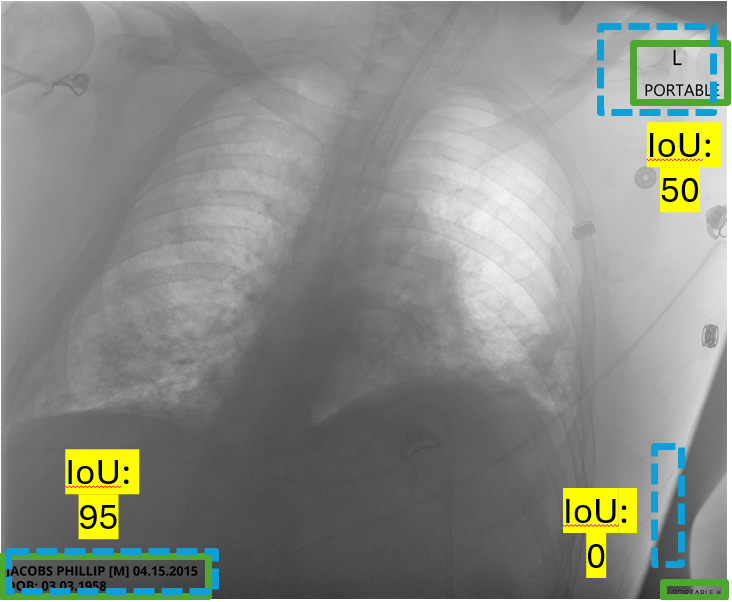}
    \caption{Examples of bounding-box overlap where green (solid line) boxes are the labels and blue (dashed line) are the predictions: bottom left shows a high IoU of 95\%, top right shows an allowable IoU of 50\%, and bottom right is a IoU of 0\%.}
    \label{fig:iou_examples}
\end{figure}

Mathematically, it is defined as:
\begin{equation}
\text{mAP}_{50} = \frac{1}{N_{\text{classes}}} \sum_{c=1}^{N_{\text{classes}}} \text{AP}_{50}^c
\label{eq:map}
\end{equation}

where \(\text{AP}_{50}^c\) is the average precision for class \(c\) calculated by measuring the area under the Precision-Recall curve at 50\% IoU. We then average these values over all \(N_{\text{classes}}\) classes. The IoU metric, which measures the overlap between predicted bounding boxes and ground-truth bounding boxes, is particularly relevant for text-detection tasks because burned-in text can appear at varying resolutions, orientations, or positions, as illistrated in Figure~\ref{fig:iou_examples}. An IoU threshold of 0.5 balances penalizing predictions that are far from the ground-truth while still allowing minor offsets. An IoU of 0.5 is widely accepted because it ensures that the recognized bounding box covers most of the text region, while not overly penalizing small shifts or shape variations inherent in real-world medical images.

\begin{figure}[h]
    \centering
    \includegraphics[width=\textwidth]{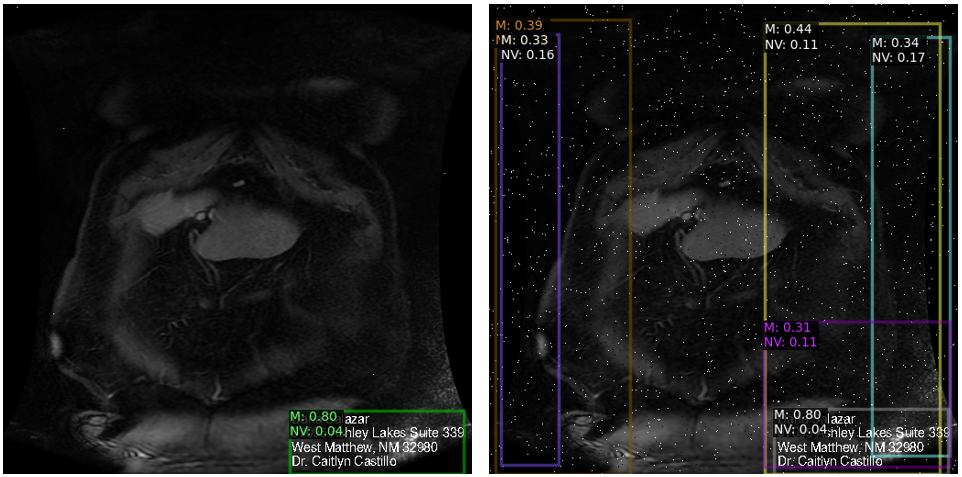}
    \caption{Examples of the effects of salt and pepper noise on model predictions (class prediction mean, M) and the uncertainty metric (normalized variance, NV). Left image: minimal noise condition (SNR = 124); right image: high noise condition (SNR = 8). Note that the uncertainty metric (NV) increases significantly for all false-positive predictions under noisy conditions, indicating the model's sensitivity to degraded image quality.}
    \label{fig:salt and pepper}
\end{figure}

For the second component, we introduce noise resilience testing by adding salt and pepper noise to the images at varying signal-to-noise ratios (SNR):
\begin{equation}
\text{SNR} \in \{124, 90, 64, 45, 32, 23, 16, 14, 12, 8, 6, 4, 2\}
\label{eq:snr_values}
\end{equation}
where \(\text{SNR} = 124\) corresponds to no noise and \(\text{SNR} = 2\) represents extreme noise, as seen in Figure \ref{fig:salt and pepper}. 

For each SNR level, we calculate the mean normalized uncertainty statistic \(\mbox{M}_{\text{uncert}}(s)\) across a subset of 1,080 images:
\begin{equation}
\mbox{M}_{\text{uncert}}(s) = \frac{1}{N_{\text{img}}} \sum_{i=1}^{N_{\text{img}}} \frac{1}{N_{\text{det}}^i} \sum_{j=1}^{N_{\text{det}}^i} \mathcal{U}(d_{i,j}, s)
\label{eq:mean_uncertainty}
\end{equation}
where \(\mathcal{U}(d_{i,j}, s)\) is the uncertainty value for detection \(j\) in image \(i\) at SNR level \(s\), \(N_{\text{img}}\) is the number of images, and \(N_{\text{det}}^i\) is the number of detections in image \(i\).

\begin{figure}[h]
    \centering
    \includegraphics[width=0.7\textwidth]{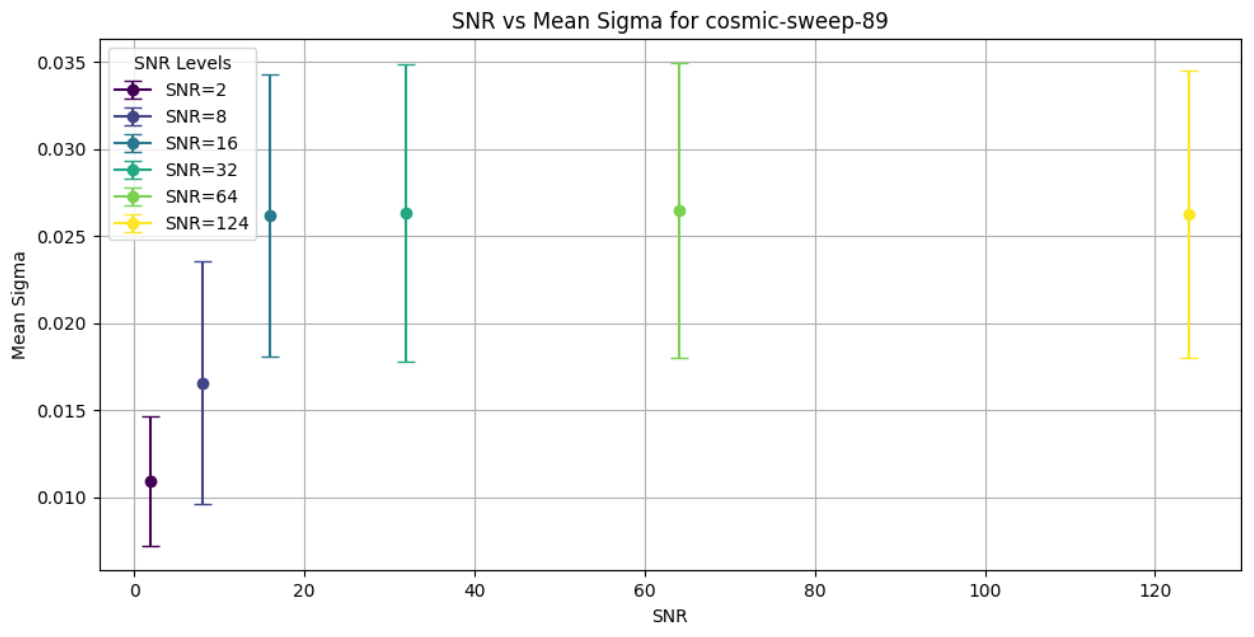}
    \vspace{0.5em} 
    \includegraphics[width=0.7\textwidth]{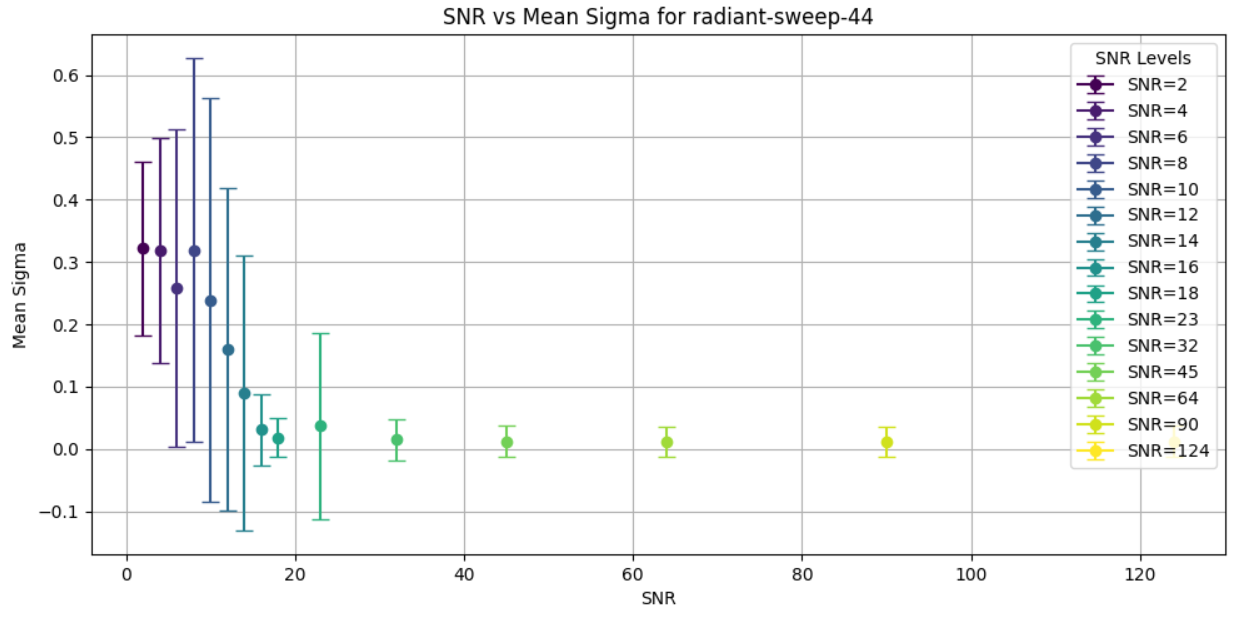}
    
    \caption{Examples illustrating the effect of loss coefficients on a model's reactance to out-of-distribution data. \textbf{Top:} A model with poor reactance, where the mean uncertainty metric (mean sigma) counterintuitively decreases as noise increases (lower SNR values indicate higher noise). \textbf{Bottom:} A model demonstrating good reactance, characterized by a clear negative slope—uncertainty increases appropriately with rising noise—and minimal variability (small error bars) under negligible noise conditions (SNR $>$ 30 dB).}
    \label{fig:reactance}
\end{figure}

We then calculate the slope \(S_{\text{noise}}\) of the uncertainty response to decreasing SNR:
\begin{equation}
S_{\text{noise}} =  \frac{d\mbox{M}_{\text{uncert}}(s)}{d(\text{SNR}^{-1})} 
\label{eq:noise_slope}
\end{equation}
A negative slope indicates that the mean uncertainty metric is higher under high noise than under low noise—that is, we aim for a more pronounced negative slope. This slope constitutes our second objective function term, representing the model’s reactance to noise, as shown in Figure \ref{fig:reactance}.

For the third term, we analyze the variance of the uncertainty statistic under low noise conditions (\(\text{SNR} > 30\)):
\begin{equation}
V_{clean} = \frac{1}{|\mathcal{S}_{\text{clean}}|} \sum_{s \in \mathcal{S}_{\text{clean}}} \text{Var}(\mathcal{U}(s))
\label{eq:variance_clean}
\end{equation}
where \(\mathcal{S}_{\text{clean}} = \{s \in \text{SNR} \mid s > 30\}\) and \(\text{Var}(\mathcal{U}(s))\) is the variance of uncertainty values at SNR level \(s\). This metric quantifies consistency in uncertainty estimation when minimal noise is present, as seen in the error bars of Figure \ref{fig:reactance}; the top image has large error bars indicating high variance in uncertainty metric especally when little noise is present $\text{SNR} > 30$dB, while the bottom image has tight error bars indicating consistent uncertainty esitmation under low noise (in distribution data).

\begin{figure}[h]
\centering
\includegraphics[width=0.7\linewidth]{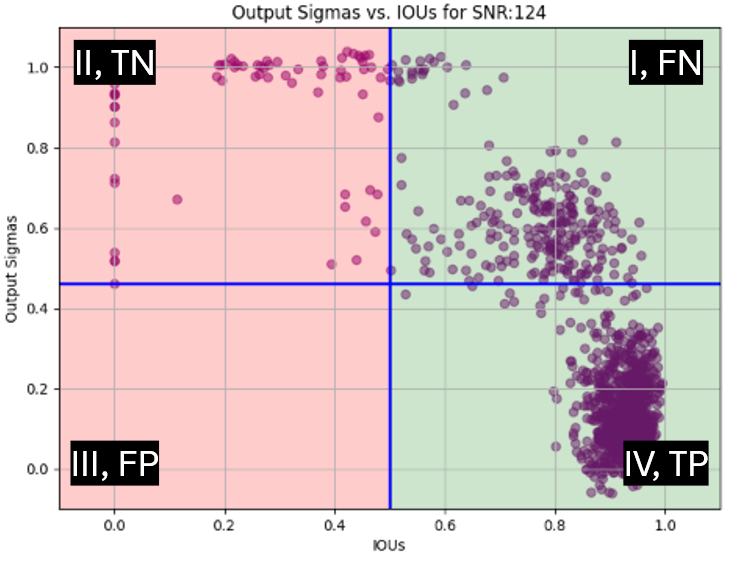}
\caption{Relationship between uncertainty and IoU scores. The plot is divided into four quadrants: (Q1, Top-right) False Negatives, representing correct detections but with high uncertainty; (Q2, Top-left) True Negatives, indicating incorrect detections appropriately flagged with high uncertainty; (Q3, Bottom-left) False Positives, incorrect predictions misleadingly indicated by low uncertainty; and (Q4, Bottom-right) True Positives, correct predictions with appropriately low uncertainty.}
\label{fig:output_sigma_vs_iou}
\end{figure}

The fourth and fifth terms pertain to uncertainty thresholding. We plot the uncertainty metric against the Intersection over Union (IoU) to establish a clear decision boundary, defined by:

\begin{equation}
\text{IoU}_{\text{thresh}} = \max_{u} \{u \mid \text{FP}(u) = 0\}
\label{eq:iou_threshold}
\end{equation}

where \(\text{FP}(u)\) is the number of false positives when using uncertainty threshold \(u\). This threshold identifies the maximum uncertainty level at which no false positives occur, considering detections with  \(\text{IoU} < 0.5\) as incorrect. This boundary is depicted by the blue horizontal line in Figure \ref{fig:output_sigma_vs_iou}.

The false negative rate based on this threshold is then calculated as:

\begin{equation}
\text{FNR}_{\text{thresh}} = \frac{\text{False Negatives}}{\text{Total Predictions}} = \frac{\text{FN}}{\text{TP} + \text{FP} + \text{TN} + \text{FN}}
\label{eq:false_negatives}
\end{equation}

Here, TP (True Positives) refers to correct predictions with low uncertainty, FN (False Negatives) to correct detections with misleadingly high uncertainty, TN (True Negatives) to incorrect detections correctly identified by high uncertainty, and FP (False Positives) to incorrect predictions incorrectly indicated by low uncertainty. Minimizing the false negative rate ensures better reliability and reduces the risk of overlooked detections.

With the objective function \(\Psi\) defined in Equation~\ref{eq:objective}, we employed Bayesian optimization to tune our hyperparameters efficiently. Let 
\[
\theta \in \Theta
\]
represent the set of hyperparameters to be optimized (defined in Equation~\ref{eq:total_loss_line}). For each candidate \(\theta\), the model is trained and evaluated according to Equation \ref{eq:objective}. Bayesian optimization constructs a probabilistic surrogate model \(p(\Psi(\theta)\mid \mathcal{D})\) for the objective function, where \(\mathcal{D}\) is the set of all \((\theta,\Psi(\theta))\) pairs evaluated so far. Commonly, this surrogate is chosen to be a Gaussian Process (GP), although other models (e.g., random forests) can be used:
\[
p\bigl(\Psi(\theta)\;\bigm|\;\mathcal{D}\bigr).
\]

An acquisition function \(\mathcal{A}(\theta)\) then determines the next hyperparameter configuration to explore by balancing exploration of uncertain regions and exploitation of regions with promising \(\Psi\)-values. Typical choices include Expected Improvement (EI), Probability of Improvement (PI), or Upper Confidence Bound (UCB). In the case of EI, for example, \(\mathcal{A}(\theta)\) can be written as:
\[
\mathcal{A}_{\text{EI}}(\theta) 
= \mathbb{E}\bigl[\max\{\Psi(\theta)-\Psi^*,\,0\}\bigr],
\]
where \(\Psi^*=\max_{\theta'\in\mathcal{D}}\,\Psi(\theta')\) is the best value observed so far. The next point to evaluate is
\[
\theta_{\text{next}} 
= \arg\max_{\theta\in\Theta}\,\mathcal{A}(\theta).
\]

In practice, we used Weights \& Biases to implement this Bayesian hyperparameter sweep, systematically exploring the parameter space based on the surrogate model and acquisition function. This approach efficiently converges toward hyperparameter configurations \(\theta\) that maximize \(\Psi(\theta)\), without requiring exhaustive search over all possible combinations. By automating the selection of the next hyperparameter set, Bayesian optimization enables us to focus computational resources on promising regions, ultimately yielding stronger reactance-to-noise, improved bounding-box precision, and more consistent uncertainty estimation in our de-identification framework.

\section{Results}
\subsection{Named Entity Recognition (NER) Performance}
We evaluated the LUKE NER model on our synthetic admission notes and tag-value metadata pairs. The model showed strong zero-shot performance initially. However, to refine accuracy for our specialized PHI classes while minimizing overfitting or catastrophic forgetting, we fine-tuned for only two epochs, keeping all other training parameters at their default settings.

Table~\ref{tab:ner_training_luke} presents LUKE’s training and validation metrics over two epochs. Even with limited fine-tuning, the model showed consistent improvements in both precision and recall, resulting in robust F1 and accuracy scores. 

\begin{table}[h!]
\centering
\caption{Training and validation metrics for the LUKE model over two fine-tuning epochs.}
\label{tab:ner_training_luke}
\begin{tabular}{r|c|c|c|c|c|c}
\hline
\textbf{Epoch} & \makecell{\textbf{Training} \\ \textbf{Loss}} & \makecell{\textbf{Validation} \\ \textbf{Loss}} & \textbf{Precision} & \textbf{Recall} & \textbf{F1} & \textbf{Accuracy} \\
\hline
1 & 0.02970 & 0.01804 & 0.89836 & 0.91657 & 0.90737 & 0.99580 \\
2 & 0.01370 & 0.01405 & 0.92802 & 0.93996 & 0.93395 & 0.99695 \\
\hline
\end{tabular}
\end{table}

Despite having only two epochs of fine-tuning, LUKE achieved strong detection accuracy across multiple PHI categories. This outcome suggests that brief training can be effective when initial zero-shot performance is already high.

\subsection{Faster-RCNN with Variational Density Propagation Classification Head Training Results}

In Section \ref{sec: data for pixeldata}, we generated 10,000 images consisting of CT, MRI, and X-ray modalities with injected PHI. This dataset was initially split into 8,000 training and 2,000 testing images to determine an optimal set of loss coefficients for our model, as described in Equation \ref{eq:total_loss_line}. For our experiments, we fixed the learning rate at 0.1 (the default Faster-RCNN learning rate), set the batch size to 8, and trained for 369 epochs. The optimization process varied several hyperparameters, including the gradient clip threshold and all loss coefficients (\(\lambda_{i}\)). This approach allowed us to systematically explore the hyperparameter space to maximize our objective function while minimizing the number of training runs required to identify optimal parameter settings. All experiments were conducted on a Tesla V100-SXM2-16GB GPU (NVIDIA-SMI 550.120, Driver Version 550.120, CUDA Version 12.4) with 16 GB of GPU memory, and we allowed the Bayesian sweep to run for 177 configurations.

After determining the optimal loss coefficients (which are withheld as proprietary) using our custom data, we set up a larger training process by incorporating all available data. Specifically, we folded our 2,000 test images into the training set and combined them with 1,693 images from the MIDI-B challenge training set. To enhance data variability, we applied several augmentation techniques—such as pixel inversion, random padding, horizontal/vertical flips, and rotations within a ±30° range. We also introduced both salt-and-pepper and Gaussian noise, as well as random or distribution-based resizing to diversify image dimensions. This suite of augmentations yielded a substantially more varied dataset to bolster the robustness of our model training. The end result was a training set comprising 36,220 images, with 26,220 of them containing objects, totaling 51,896 objects in the dataset; as compared to soley training on MIDI train set of 1,693 images, where only 14 images have objects. The validation was then carried out on the 23,841 images designated as the validation set in the MIDI Challenge.

\begin{table}[h!]
\centering
\begin{tabular}{l c c}
\hline
\textbf{Metric} & \textbf{IoU} & \textbf{Value} \\
\hline
mAP & 0.50:0.95 & 0.779 \\
mAP & 0.50 & 0.997 \\
mAP & 0.75 & 0.885 \\
\hline
\end{tabular}
\caption{Mean Average Precision (mAP) of Uncertainty Aware Faster RCNN at different IoU thresholds on MIDI Challenge Validation Set.}
\label{tab:performance}
\end{table}

As shown in Table \ref{tab:performance}, the model achieves a mean Average Precision (AP) of 0.779 across IoU thresholds ranging from 0.50 to 0.95, with near-perfect performance at IoU = 0.50 (AP = 0.997). These results highlight both the strong detection capabilities of the model and its robustness across varying object boundary conditions. However, no uncertainty-metrics reactance could be calculated on this set, since the salt-and-pepper noise images were included in the training set, making them no longer out-of-distribution data.

\subsection{Full Framework Results}
Table~\ref{tab:standards performance} summarizes the performance of our de-identification framework across multiple datasets and categories. Each category represents a specific set of DICOM tags, HIPAA compliance rules, or standards from the Cancer Imaging Archive (TCIA).

\begin{table}[ht]
\centering
\caption{Results of De-identification Framework Performance across DICOM, HIPAA, and TCIA Standards}
\label{tab:standards performance}
\begin{tabular}{llllr}
\toprule
\textbf{category} & \textbf{subcategory} & \textbf{Fail} & \textbf{Pass} & \textbf{Total} \ \% \\
\midrule
\multirow{4}{*}{dicom} & DICOM-IOD-1 & 112 & 170534 & 170646 (0.9993) \\
 & DICOM-IOD-2 & 13 & 36354 & 36367 (0.9996) \\
 & DICOM-P15-BASIC-C & 0 & 429 & 429 (1.0000) \\
 & DICOM-P15-BASIC-U & 0 & 40633 & 40633 (1.0000) \\
\midrule
\multirow{7}{*}{hipaa} & HIPAA-A & 2 & 674 & 676 (0.9970) \\
 & HIPAA-B & 2 & 30 & 32 (0.9375) \\
 & HIPAA-C & 2 & 2816 & 2818 (0.9993) \\
 & HIPAA-D & 0 & 20 & 20 (1.0000) \\
 & HIPAA-G & 0 & 149 & 149 (1.0000) \\
 & HIPAA-H & 0 & 658 & 658 (1.0000) \\
 & HIPAA-R & 0 & 41578 & 41578 (1.0000) \\
\midrule
\multirow{11}{*}{tcia} & TCIA-P15-BASIC-D & 0 & 198 & 198 (1.0000) \\
 & TCIA-P15-BASIC-X & 0 & 13 & 13 (1.0000) \\
 & TCIA-P15-BASIC-X/Z/D & 0 & 147 & 147 (1.0000) \\
 & TCIA-P15-BASIC-Z & 0 & 482 & 482 (1.0000) \\
 & TCIA-P15-BASIC-Z/D & 0 & 11 & 11 (1.0000) \\
 & TCIA-P15-DESC-C & 44 & 2304 & 2348 (0.9813) \\
 & TCIA-P15-DEV-C & 0 & 17 & 17 (1.0000) \\
 & TCIA-P15-DEV-K & 2 & 177 & 179 (0.9888) \\
 & TCIA-P15-MOD-C & 0 & 2565 & 2565 (1.0000) \\
 & TCIA-P15-PAT-K & 0 & 1113 & 1113 (1.0000) \\
 & TCIA-P15-PIX-K & 0 & 29471 & 29471 (1.0000) \\
 & TCIA-PTKB-K & 117 & 31670 & 31787 (0.9963) \\
 & TCIA-PTKB-X & 257 & 901 & 1158 (0.7781) \\
 & TCIA-REV & 119 & 217651 & 217770 (0.9995) \\
\midrule
\multicolumn{2}{l}{\textbf{Total}} & 670 & 580595 & 581265 (0.9988) \\
\bottomrule
\end{tabular}
\end{table}

The results in Table~\ref{tab:standards performance} showcase the number of passes and failures for each sub-category, along with the total number of instances and the pass rate percentage. Our framework demonstrated high accuracy across all categories, with an overall success rate of 99.88\%. Notably, the DICOM and HIPAA components showed near-perfect results, ensuring comprehensive and reliable de-identification of medical images. The few observed failures were concentrated in a small subset of the TCIA subcategories, which we are addressing in ongoing work to improve generalization in more diverse image contexts.

\section{Discussion}
In this paper, we present a comprehensive de-identification framework designed to ensure the privacy and security of medical imaging data while preserving its utility for research and clinical applications. Although high-performance automation is essential, truly secure and scalable data sharing requires more than just accuracy—it demands explainability, auditability, and the ability to manage residual risk in production-grade, compliance-sensitive environments. Our approach achieves these goals by integrating rule-based logic with advanced AI-driven techniques, further enhanced through a novel uncertainty quantification pipeline.

A key strength of our framework lies in its flexibility and user-friendliness. We have developed a sophisticated User Interface (UI) that enables users to go beyond simple redaction; they can replace, blank, or retain specific data elements depending on downstream needs. This UI also supports longitudinal date adjustments, ensuring consistent shifts across an entire dataset while preserving temporal relationships. In addition, users can readily review and validate precisely which DICOM tags have been removed, facilitating both trust and transparency.

From a modeling perspective, our system incorporates a Faster R-CNN architecture with variational density propagation to estimate the confidence of text-region detections in pixel data. This uncertainty-aware design allows us to adopt a risk-aware redaction process in which high-certainty detections are processed automatically, while uncertain cases trigger a manual review. The approach not only improves reliability but also introduces features such as “Expert Determination readiness,” enabling institutions to adjust thresholds and audit settings to meet specific legal or organizational standards. Furthermore, ambiguous instances can be quarantined, allowing the broader dataset to be processed without delays while uncertain items await further adjudication.

To address the well-known challenges of data scarcity and limited access to diverse, real-world PHI examples, we have used synthetic datasets for training robust models. However, synthetic data alone may not capture all real-world complexities. Consequently, we have initiated a formal, IRB-approved retrospective clinical study at Moffitt Cancer Center to validate our framework using real patient imaging data. Feedback from clinicians and operational experts has been vital for refining our models, confirming that our approach performs robustly in practice—not just in controlled simulations.

Looking ahead, we intend to expand uncertainty estimation to both the bounding box regression module and metadata-based NLP components. By evaluating all candidate bounding boxes—not just the final predictions—we can further mitigate the risk of overlooked PHI and enable more granular explanations. Applying similar uncertainty-aware strategies to Named Entity Recognition in metadata will facilitate consistent and transparent risk management across text-based tags as well. Beyond DICOM, we are also interested in extending our methodology to meet international privacy regulations (e.g., GDPR) and to handle multimodal data sources, such as clinical notes, pathology reports, and omics data.

In conclusion, our de-identification framework combines the transparency and configurability of traditional methods with the adaptability and depth of state-of-the-art AI techniques. By integrating a user-friendly interface, calibrated uncertainty modeling, and real-world validation, we provide a risk-calibrated, auditable, and highly configurable solution that bridges the gap between research innovation and operational trust. Our work not only protects patient privacy but also unlocks the full potential of real-world imaging data for continued advances in healthcare research.

\section*{Acknowledgments}
This work was supported in part by the National Cancer Institute (NCI) under a Phase I SBIR contract awarded to Impact Business Information Solutions (IBIS). We thank our collaborators at Moffitt Cancer Center for their invaluable partnership, including clinical and operational stakeholders who contributed critical insights through a formal IRB-approved validation study. We are especially grateful to Dr. Ghulam Rasool and the Moffitt informatics team for enabling real-world validation and iterative refinement of our models.
We also acknowledge the contributions of Nidhal Bouaynaya and the team at Rowan University for their expertise in uncertainty modeling and computer vision, and the broader MIDI-B Challenge community for fostering a high-impact venue for benchmarking and collaboration.

\section*{Disclosure of Interests}
The authors declare no competing financial interests. Some authors are affiliated with Impact Business Information Solutions (IBIS), the primary developer of the de-identification framework described in this manuscript.


\acks{We thank the entire IBIS team and our collaborators at Moffitt Cancer Center and Rowan University for their support and contributions to this work.}

\ethics{This study was conducted in accordance with all applicable regulations concerning human subjects research. The retrospective validation study performed at Moffitt Cancer Center was approved by their Institutional Review Board (IRB) and adhered to ethical standards governing data privacy and patient protection. All synthetic data used for model development and pre-validation were generated to avoid the use of real PHI.}

\coi{The authors have no conflicts of interest to report.}

\data{Synthetic datasets generated for this study can be made available upon reasonable request. The use of real clinical data from Moffitt Cancer Center was conducted under IRB-approved protocols and is not publicly available due to privacy restrictions.}

\bibliographystyle{plainnat}

\end{document}